\pdfoutput=1
\documentclass[10pt,logo,copyright]{nvidiatechreport}
\usepackage[numbers,sort]{natbib}
\usepackage[utf8]{inputenc} %
\usepackage[T1]{fontenc}    %
\usepackage{xcolor}
\definecolor{cvprblue}{rgb}{0.21,0.49,0.74}
\usepackage{url}            %
\usepackage{booktabs}       %
\usepackage{amsfonts}       %
\usepackage{nicefrac}       %
\usepackage{microtype}      %

\usepackage{tabularx}

\usepackage{graphicx}
\usepackage{graphics} %
\usepackage{epsfig} %
\usepackage{amsmath} %
\usepackage{amssymb}  %
\usepackage{enumitem}
\usepackage{mathtools} 
\usepackage{algorithm}
\usepackage{algorithmic}
\usepackage{wrapfig}
\usepackage{colortbl}
\usepackage{listings}
\usepackage{afterpage}
\usepackage{relsize}
\usepackage{multicol}
\usepackage{multirow}
\usepackage{wasysym}
\usepackage{caption}
\usepackage{subcaption}
\usepackage{sidecap}
\usepackage{pifont}
\usepackage{makecell}    %
\usepackage{diagbox}
\usepackage{url}
\usepackage[symbol]{footmisc}
\usepackage{cleveref}
\usepackage{dsfont}
\usepackage{xspace}

\usepackage{tikz}
\definecolor{catBlue}{HTML}{3D78B8}
\definecolor{catOrange}{HTML}{EA8E36}
\definecolor{catTeal}{HTML}{49A39A}
\definecolor{catRed}{HTML}{D25455}
\definecolor{catGold}{HTML}{D8AB2A}
\definecolor{catGreen}{HTML}{569C45}

\newcommand{\colordot}[1]{\tikz[baseline=-0.55ex]\fill[#1] circle (0.5ex);}

\newcommand{\name}{Vesta\xspace}

\title{Vesta: A Generalist Embodied Reasoning Model}

\author{Johan Bjorck$^{*}$, ~~Zhiqi Li$^{*}$, ~~Yunze Man$^{*1}$, ~~Jing Wang$^{*}$, ~~An-Chieh Cheng$^{\dagger2}$, ~~Sifei Liu$^{\dagger}$, ~~Shihao Wang$^{\dagger3}$, ~~Zhiding Yu$^{\dagger}$, ~~Abhishek Badki, ~~Stan Birchfield, ~~Valts Blukis, ~~Yevgen Chebotar, ~~Siyi Chen$^{4}$, ~~Sicong Leng$^{5}$, ~~Yu-Cheng Chou$^{6}$, ~~Tianli Ding, ~~Boyi Li, ~~Zhengyi Luo, ~~Hang Su, ~~Jonathan Tremblay, ~~Tingwu Wang, ~~Bowen Wen, ~~Jimmy Wu, ~~Xianghui Xie$^{7}$, ~~Hanrong Ye, ~~Hongxu Yin, ~~K.R. Zentner, ~~Liangyan Gui$^{1}$, ~~Yu-Xiong Wang$^{1}$, ~~Yuke Zhu$^{\ddagger}$, ~~Linxi "Jim" Fan$^{\ddagger}$, ~~Jan Kautz$^{\ddagger}$}

\begin{document}

\begin{abstract}
Robots operating in open-world environments must seamlessly integrate localization, spatial reasoning, navigation, and long-horizon planning. While specialist models excel at individual tasks, deploying a multi-model stack is computationally expensive and prone to cascading errors. We present \name, a unified embodied generalist that consolidates these capabilities into a single foundation model. Our approach combines a diverse and massive curated corpus designed to induce spatial grounding and a simple multimodal memory harness that enables reasoning over extended time horizons. Across diverse benchmarks, \name on average beats individual SOTA baselines by >$20\%$ and beats an ensemble of per-category-best baselines by $>10\%$ -- thus demonstrating that a generalist model can match or exceed specialists. On real-world robotic tasks requiring memory and reasoning, \name improves task success by >35\%. Our work thus demonstrates that a single generalist is a feasible, scalable, and arguably preferable alternative to combining specialists.
\end{abstract}

\maketitle

\abscontent

\begin{figure}[b!]
    \centering
    \vspace{-3mm}
    \includegraphics[width=\linewidth]{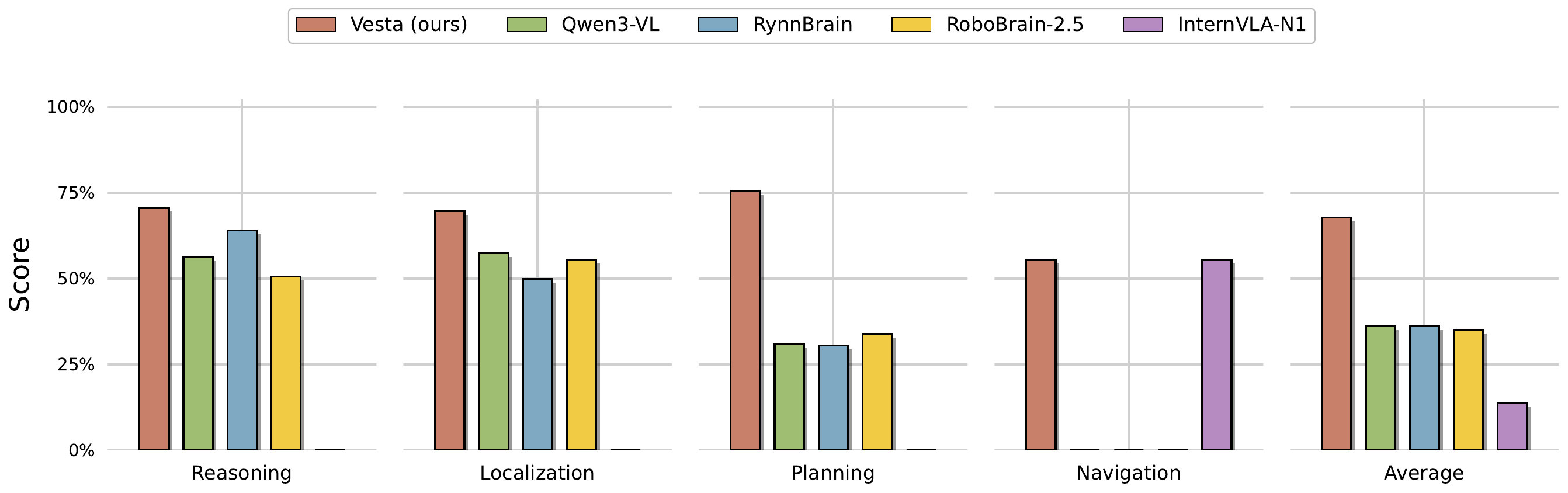}
    \caption{\name unifies localization, navigation, embodied reasoning, and action planning into a single generalist model. It scores over $20$ points above the average prior baseline and $>10$ points above the strongest baseline in each individual category. On real robots, \name~improves success by $38.3\%$ on memory-heavy tasks.}
    \label{fig:teaser}
\end{figure}

\section{Introduction}
\label{sec:intro}

Robots operating in the real world must bridge the gap between high-level semantic reasoning and low-level physical execution. Consider a humanoid robot cleaning a grocery store: it must simultaneously master complex motor skills (\textit{e.g.}, scrubbing floors) and sophisticated logic (\textit{e.g.}, distinguishing trash from misplaced products or correctly answering questions from shoppers that approach the robot). An emerging paradigm is to decouple these demands into a hierarchical stack: a planner Vision-Language Model (VLM) generates high-level instructions, which are then executed by a specialized action VLA model~\citep{shi2025hirobot,arxiv2025gpc}. Under this framework, the planner VLM commonly handles diverse skills such as spatial localization \citep{ji2025robobrain}, memory \citep{memer2026}, navigation skills \citep{cheng2024navila}, world knowledge and more.

The academic literature typically treats these capabilities as isolated challenges. Specialist models are often developed in silos: navigation specialists are optimized for navigation simulation benchmarks~\citep{cheng2024navila, rxr}, memory specialists are focused on long-horizon manipulation~\citep{memer2026,mem2026}, and reasoning specialists are optimized for static question-answering suites~\citep{dang2026rynnbrain,tan2026robobrain}. While these specialists perform well in-domain, deploying an ensemble with one specialist per capability is not scalable. Such modularity introduces latency, complicates the inference stack, and is prone to cascading failures where an error in one specialist's output propagates through the system.

\begin{figure}[t!]
    \centering
    \vspace{-3mm}
    \includegraphics[width=\linewidth]{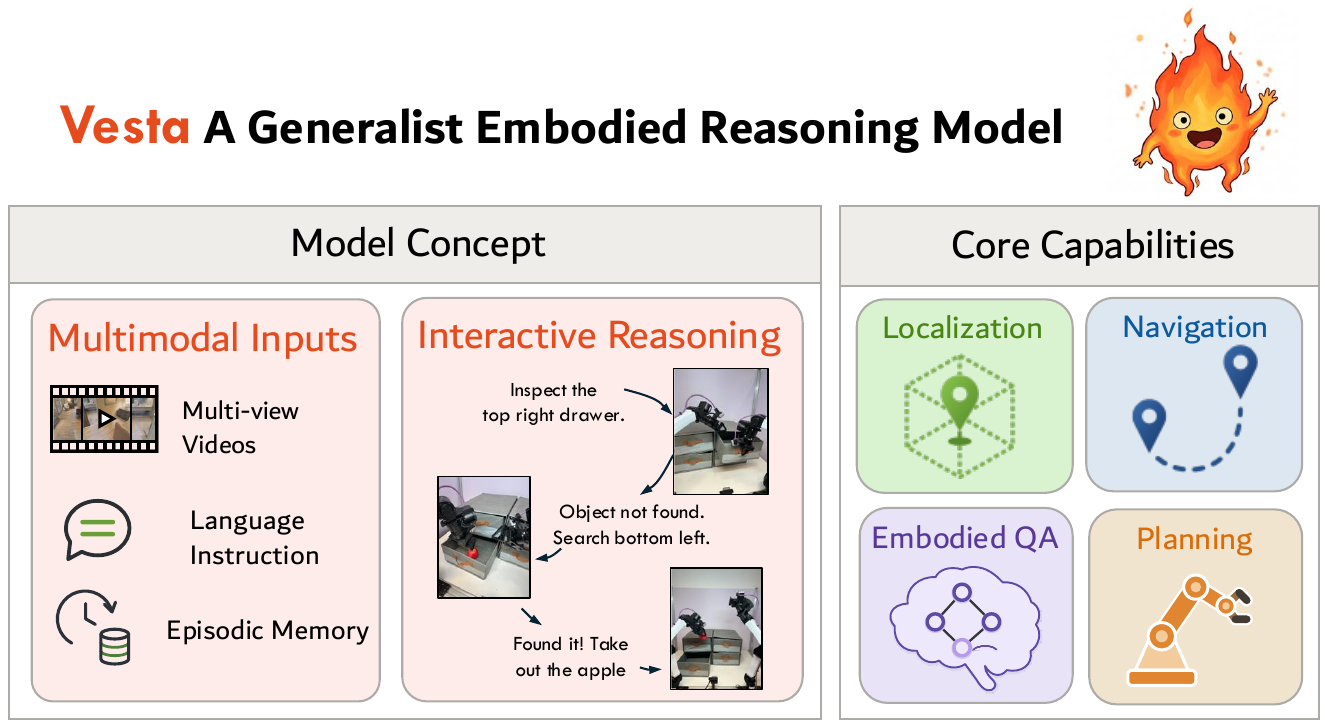}
    \caption{\name is a generalist embodied model, supporting multimodal inputs and hierarchal control. The four main capabilities are localization, navigation embodied question-answering and real-world planning.}
    \label{fig:overview}
\end{figure}

We instead posit that these capabilities can—and should—be unified into a single generalist planner. To this end we present \textbf{\name}, a generalist embodied model built around three core techniques. Firstly, a \textit{curated supervised fine-tuning (SFT) corpus} covering grounding, navigation, embodied reasoning, and real-robot data, targeted toward spatially grounded capabilities. Secondly, a \textit{simple multimodal memory harness} that interleaves history image frames with a running textual cache of past subtasks. 
Empirically, \name~beats SOTA baselines across diverse benchmarks (see \Cref{fig:teaser}). Across the four capabilities we evaluate, \name~ on average scores >$20\%$ above the strongest single baseline and $>10\%$ over an oracle ensemble of baselines (i.e., using the best baseline in each individual category). This demonstrates that it's possible to unify these capabilities into one generalist without hurting benchmark scores. When deployed on a real robot platform, \name~improves average task success by $38.3\%$ on tasks that require memory and long-horizon reasoning. 

Our contributions are (1) \name, a generalist embodied model that matches or exceeds the performance of domain specialists across four distinct capability axes. (2) A training recipe combining a simple memory harness with a diverse SFT-corpus that yields superior cross-task generalization compared to naive supervised fine-tuning. (3) Empirical validation on a bimanual robot platform demonstrating that \name significantly improves the execution of long-horizon, memory-intensive tasks. (4) Taken together, our work demonstrates that generalist planners are a feasible, scalable, and, we argue, preferable alternative to combining specialists.

\begin{figure}
    \centering
    \vspace{-3mm}
    \includegraphics[width=\linewidth]{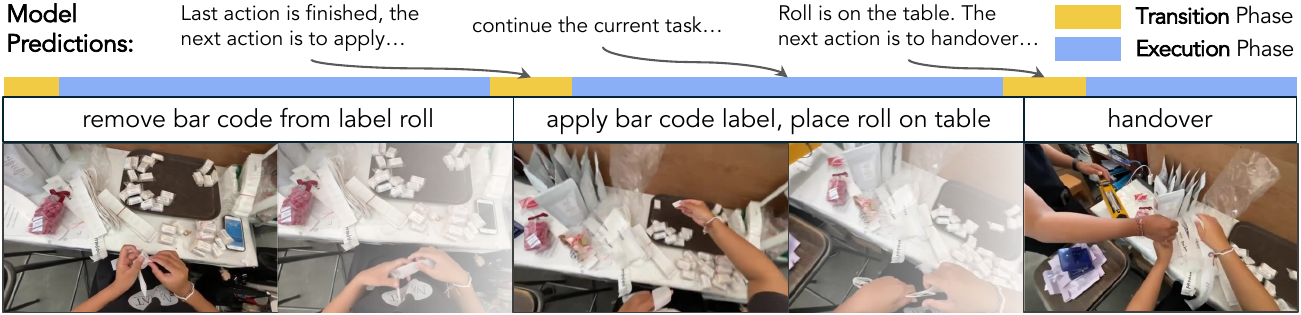}
    \caption{\textbf{Demonstration of the action planning task}. The model is tasked to plan the next subtask based on the overall objective and the memory context. Intermediate steps are omitted.}
    \label{fig:nextaction_demonstration}
\end{figure}

\section{Methods}

\label{sec:methods}

\name is finetuned from the Qwen3-VL-8B \citep{bai2025qwen3} base model. 
Our supervised fine-tuning (SFT) strategy builds base capabilities in localization, navigation, embodied reasoning, and memory-conditioned planning. 
Significant effort has gone into data curation, which is detailed below.
\subsection{Localization}

\name is endowed with grounding capabilities. It can associate text descriptions with spatial regions and perform pointing to predict contact or manipulation points. Together, these capabilities serve as the planner's interface between perception and action. We design the localization dataset using a base–tail strategy. The base component uses large-scale grounding and detection datasets: Objects365~\cite{Shao2019Objects365}, COCO~\cite{Lin2014COCO}, and LVIS~\cite{Gupta2019LVIS}. These provide broad category coverage and dense annotations, establishing general-purpose grounding priors. The tail component adds embodied and robotics-specific data. This includes egocentric observations, manipulation-centric annotations, and temporally evolving interaction sequences~\cite{ji2025robobrain,huang2024manipvqa}. This component adapts the model to partial observability, viewpoint changes, and action relevance. All structured outputs, such as points and boxes, are decoded as text tokens through the same language head used for free-form answers.

\subsection{Navigation} 

Vision-and-Language Navigation (VLN) is an instruction-guided navigation task where the agent must produce navigation actions given a text instruction and egocentric observations. We adopt the standard R2R-style VLN formulation: an episode $e = (I, s_0, g)$ consists of a route instruction $I$, an initial pose $s_0$, and a target goal $g \in \mathbb{R}^3$. At each decision point $t$, the agent observes its pose $s_t$, egocentric image $o_t$, and a sampled visual history $H_t = (o_{k_0}, \ldots, o_{k_{N-1}})$. Conditioned on $(I, H_t, o_t)$, the planner $\pi_\theta$ predicts a navigation action $a_t \in \mathcal{A}$, such as pixel goal, turn sequence, or \textsc{stop}, which is executed to produce $s_{t+1}$. The episode terminates on \textsc{stop} or step budget exhaustion. The objective is

{\normalsize
  \setlength{\abovedisplayskip}{3pt}
  \setlength{\belowdisplayskip}{-3pt}
  \setlength{\abovedisplayshortskip}{-14pt}
  \setlength{\belowdisplayshortskip}{1pt}
\begin{equation}
\max_\theta \mathbb{E}_{e \sim \mathcal{D}}
\left[\mathds{1}\{d(s_T, g) \le d_\text{succ}\}\right],
\label{eq:nav_objective}
\end{equation}}

We query the VLN model at high-level decision steps $t=0,\ldots,T-1$, while low-level motion is handled by the navigation backend, following~\citet{internnav2025}. At step $t$, the prompt contains the instruction $I$, the current view $o_t$, and up to $N$ sampled history frames $H_t=(o_{k_1},\ldots,o_{k_{N_t}})$, where $0 \le k_1 < \cdots < k_{N_t} < t$ and $N_t \le N$. The model outputs one of three action forms: a pixel goal specified by a downward-view request $\downarrow$ followed by normalized waypoint coordinates $(u,v)\in[0,1000]^2$; a turn sequence over $\{\leftarrow,\rightarrow\}$, where $\leftarrow$ and $\rightarrow$ denote yaw primitives towards left and right; or \textsc{stop}. We source VLN-CE datasets from R2R~\cite{vlnce}, RxR~\cite{rxr}, and ScaleVLN~\cite{scalevln}. Episodes are rendered in simulation~\cite{habitat19iccv,mp3d,hm3d} to obtain trajectory–instruction pairs.

\subsection{Embodied Reasoning}

Embodied reasoning extends spatial localization to action-conditioned scene understanding~\cite{hao2026mimoembodiedxembodiedfoundationmodel,dang2026rynnbrain,song2025robospatial}. General visual understanding tasks, such as recognition, detection, and visual question answering, provide broad perceptual grounding. Embodied tasks build on this foundation, and sample tasks include affordance and placement prediction, manipulation trajectory generation as ordered waypoints, and task progress estimation from egocentric video~\cite{dang2026rynnbrain, arxiv2025tracespatial}. The training corpus mirrors this hierarchy: large-scale VQA, detection, and pointing data provide visual priors, while embodied data focuses on agent-centric interaction, often from real-robot and human manipulation trajectories. This yields a unified interface for \emph{what}, \emph{where}, \emph{how}, and \emph{when} reasoning.

\subsection{Action Planning with Memory}

We formulate long-horizon robot planning from egocentric video. Given a textual goal $g$ (\textit{e.g.}, ``unpack groceries''), the planner predicts the next subtask $a_t$ (\textit{e.g.}, ``open fridge'') in text form at each timestep $t$. This subtask is executed by a low-level actor. The problem is non-Markovian: $a_t$ depends on the full trajectory history. The policy $\pi$ conditions on all past observations and actions, i.e.: $a_t = \pi(o_{\leq t}, a_{<t}, g)$. To avoid intractable context lengths over dense histories, we approximate the state as $s_t = \Phi(o_t, \mathcal{M}_t, g)$, where $\mathcal{M}_t$ is the compressed context from a memory harness. The model produces a Chain-of-Thought before each subtask, with four phases: \textit{Observation} (``What do I see?''), \textit{Progress} (``How far has the task progressed?''), \textit{Reasoning} (``What should happen next and why?''), and \textit{Action} (the predicted subtask). The additional fields aid in reasoning, but only $a_t$ is written to memory. 

\textbf{Memory Harness.} \name uses an explicit memory harness. At each step $t$, the memory builds a curated context $\mathcal{M}_t$ from prior steps and re-injects it into the prompt. Each step is a tuple $m_i = \langle i, \tau_i, o_i, a_i, g \rangle$, where $i$ is the step index, $\tau_i$ the timestamp, $o_i$ the observation, $a_i$ the predicted subtask, and $g$ the goal. The history is $\mathcal{H}_t = \{m_1, \dots, m_{t-1}\}$. Historical images are capped at $K$ via a sampling operator $\mathcal{S}_t \subseteq \{1, \dots, t-1\}$ with $|\mathcal{S}_t| \leq K$. We use two strategies: \textit{uniform sampling} and \textit{recency-biased sampling} (exponential weighting). The first frame is always retained to preserve the initial state. We deliberately adopt a minimalist design: \Cref{sec:ablations} shows that this simple memory conditioning already leads to great planner evaluation.

\begin{figure}[!b]
\flushleft %
\vspace{-3mm}
\begin{minipage}[c]{0.45\linewidth}
\centering
\includegraphics[width=0.9\linewidth]{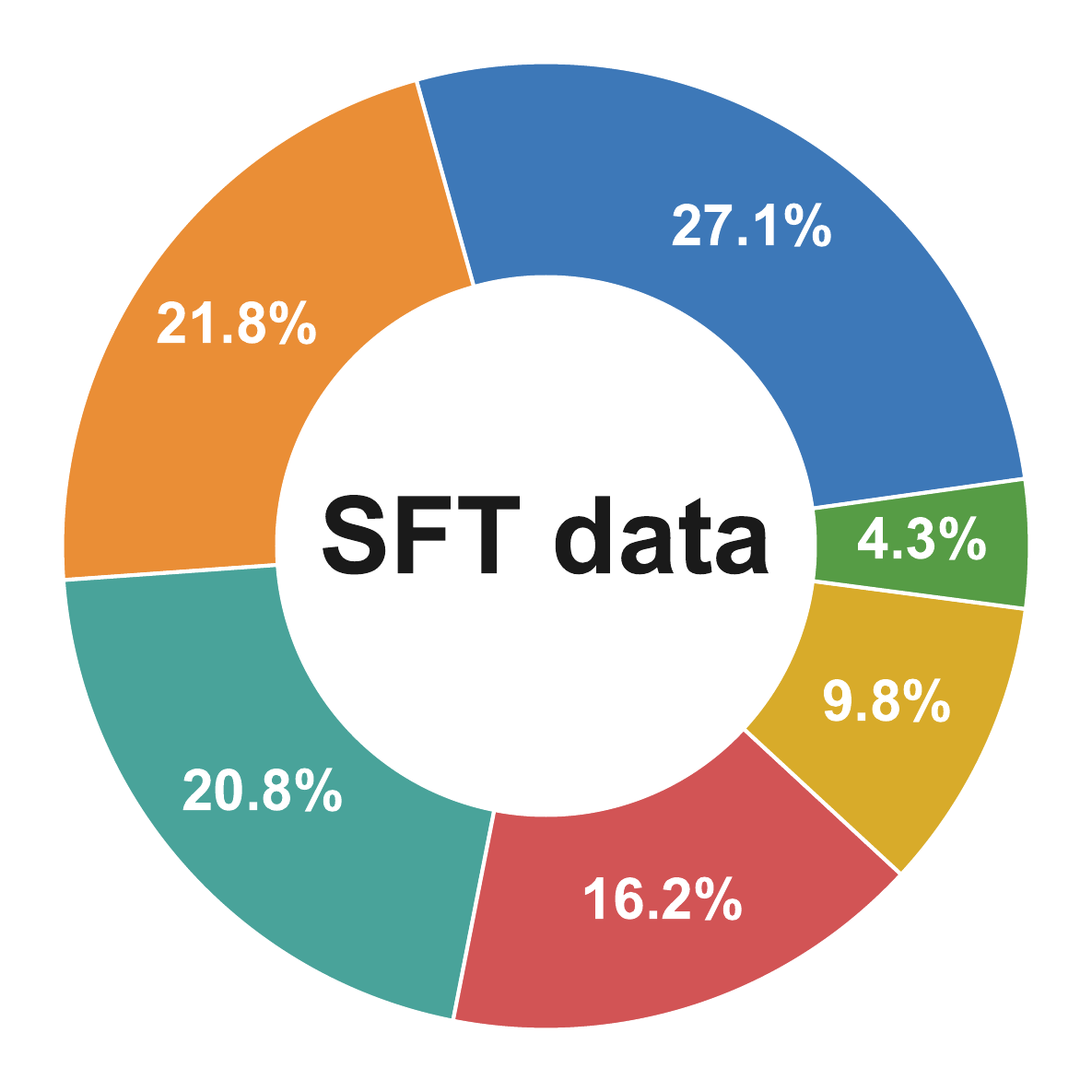}
\end{minipage}
\hfill %
\begin{minipage}[c]{0.40\linewidth} %
\centering
\renewcommand{\arraystretch}{1.2}
\resizebox{\linewidth}{!}{
\begin{tabular}{l c}
\toprule
\textbf{Category} & \textbf{Share} \\
\midrule
\colordot{catBlue}~~Spatial Intelligence & {\color{catBlue}27.1\%} \\
\colordot{catOrange}~~Navigation         & {\color{catOrange}21.8\%} \\
\colordot{catTeal}~~Grounding            & {\color{catTeal}20.8\%} \\
\colordot{catRed}~~General VLM           & {\color{catRed}16.2\%} \\
\colordot{catGold}~~Embodied Reasoning   & {\color{catGold}\phantom{0}9.8\%} \\
\colordot{catGreen}~~Real Robots         & {\color{catGreen}\phantom{0}4.3\%} \\
\midrule
\textbf{Total} & \textbf{100.0\%} \\
\bottomrule
\end{tabular}
}
\end{minipage}
\hspace{\fill} %
\vspace{-2mm}
\caption{\textbf{SFT data mixture.} Our SFT mix spans six categories.}
\label{fig:sft_data_composition}
\end{figure}

\section{Training recipe}

\label{sec:training}

Our SFT corpus is intentionally biased toward spatially grounded capabilities, see \Cref{fig:sft_data_composition}. Spatial intelligence forms the largest portion of the corpus, accounting for 27.1\% of the samples, while navigation and grounding contribute another 21.8\% and 20.8\%, respectively. General VLM data remains a substantial component at 16.2\%, serving to preserve broad visual-language competence and reduce over-specialization. The remaining embodied reasoning and real robot data provide task-level reasoning and real-world execution signals that align the model with downstream embodied settings. We train for 1 epoch over the full mixture, using a learning rate of 1e-5 and weight decay of 0.01. We train our model with 128 H100 GPUs and a batch size of 256.

\section{Evaluation}
\label{sec:evaluation}

\begin{table*}[b!]
\centering
\setlength{\tabcolsep}{8pt} 
\vspace{-3mm}
\resizebox{\linewidth}{!}{
\renewcommand{\arraystretch}{1.1}

\begin{tabular}{ll|c|ccc}
\toprule
\multicolumn{2}{c|}{\multirow{2}{*}{\diagbox[width=5cm]{\textbf{Benchmark}}{\textbf{Model}}}} & \textbf{\name} & \textbf{RynnBrain} & \textbf{RoboBrain 2.5} & \textbf{Qwen3-VL} \\ 
\multicolumn{2}{c|}{} & \textbf{8B} & \textbf{8B} & \textbf{8B} & \textbf{8B} \\
\midrule
\multirow{13}{*}{\textbf{Cognition}} 
& Open-X VQA~\cite{chen2025robo2vlm} & \textbf{89.3} & \underline{74.0} & 52.9$^\dagger$ & 59.8 \\
& SAT~\cite{ray2024sat} & \textbf{81.3} & \underline{70.0}$^\dagger$ & 67.3$^\dagger$ & 65.3$^\dagger$ \\ 
& VSI-Bench~\cite{yang2025vsibench} & \underline{64.5} & \textbf{71.0} & 42.9$^\dagger$ & 60.3$^\dagger$ \\ 
& MMSI-Bench~\cite{yang2025mmsibench} & \textbf{40.8} & \underline{39.6} & 29.4$^\dagger$ & 30.8$^\dagger$ \\ 
& ERQA~\cite{team2025erqa} & \underline{44.9} & \textbf{46.8} & 44.0$^\dagger$ & \underline{44.8} \\ 
& MindCube-Tiny~\cite{yin2025mindcube} & \textbf{80.9} & \underline{56.6} & 29.2$^\dagger$ & 36.0 \\ 
&CV-Bench~\cite{tong2024cvbench} & \textbf{88.1} & \underline{87.7}$^\dagger$& 87.6$^\dagger$ & 86.2$^\dagger$ \\
& PAI-U~\cite{zhou2025pai} & \textbf{57.9} & 56.6$^\dagger$ & 55.0$^\dagger$ & \textbf{57.9}$^\dagger$ \\ 
& EgoTaskQA~\cite{jia2022egotaskqa} & \underline{81.9} & 72.5 & \textbf{85.0}$^\dagger$ & 57.8$^\dagger$ \\  
& RoboSpatial~\cite{song2025robospatial} & 57.8  & \textbf{73.1} &  \underline{73.0} &  58.2 \\
\cmidrule{2-6}
& \textbf{Average} & \textbf{68.7} & \underline{64.8} & 56.6 & 55.7 \\
\midrule
\multirow{6}{*}{\textbf{Localization}}
& CrossPoint~\cite{wang2025crosspoint} & \textbf{76.0} & 44.3 & \underline{75.4} & 28.7 \\ 
& EmbSpatial~\cite{du2024embspatial} & \textbf{81.9} & 79.3$^\dagger$ & \underline{75.8} & 78.5$^\dagger$ \\ 
& Where2Place~\cite{yuan2024robopoint} & \textbf{68.3} & \underline{66.9}$^\dagger$ & 66.0$^\dagger$ & 64.7$^\dagger$ \\ 
& RefSpatial~\cite{zhou2025roborefer} & \underline{59.9} & 59.2 & \textbf{60.5} & 53.4 \\ 
& PointBench~\cite{cheng2025pointarena} & \underline{63.2} & 59.7$^\dagger$ & \textbf{69.1} & 61.4$^\dagger$ \\ 
\cmidrule{2-6}
& \textbf{Average} & \textbf{69.9} & 61.9 & \underline{69.4} & 57.3 \\
\bottomrule
\end{tabular}}
\vspace{-1mm}
\caption{\textbf{Embodied benchmarks.} We compare \name against SOTA baselines of the same size. Across both embodied cognition and localization, our model beats the baselines. $\dagger$ marks results obtained using our evaluation code. Note that navigation specialist like InternVLA-N1~\cite{wang2025internvla} completely fails out of domain due to catastrophic forgetting, always outputting $\rightarrow\rightarrow$ regardless of questions types.}
\label{tab:embodied_benchmark} 
\end{table*}

\subsection{Embodied Reasoning}

We report scores on embodied benchmark results in \Cref{tab:embodied_benchmark}. \name delivers strong performance across both cognition and localization benchmarks, achieving the highest average score in each category. It attains the best score on the majority of cognition benchmarks and remains competitive on the rest. \name also leads on the majority of localization benchmarks and stays within a narrow margin on the remaining one, producing a clear average improvement over the existing 8B model. This demonstrates that our data strategy delivers strong, balanced capability across both categories.

\begin{table*}
\centering 
\resizebox{\linewidth}{!}{
\renewcommand{\arraystretch}{1.1}

\begin{tabular}{l ccccc c c}
\toprule 
\multirow{2}{*}{\textbf{Model}} & \multicolumn{5}{c}{\textbf{AgiBot}} & \multicolumn{1}{c}{\textbf{Egocentric-Human}} & \multirow{2}{*}{\textbf{Avg.}} \\
\cmidrule(lr){2-6} \cmidrule(lr){7-7}
& $\mathtt{CD}$ & $\mathtt{PF}$ & $\mathtt{SP}$ & $\mathtt{FS}$ & $\mathtt{RS}$ & $\mathtt{Diverse\ \ Tasks}$ & \\
\midrule
RoboBrain-2.5-8B~\cite{tan2026robobrain} & 35.3 & 81.6 & 15.9 & 38.3 & 33.0 & 27.0 & 38.5 \\
Qwen3-VL-8B~\cite{bai2025qwen3} & 36.7 & 67.8 & 18.1 & 22.1 & 30.2 & 26.7 & 33.6 \\
RynnBrain-8B~\cite{dang2026rynnbrain} & 38.7 & 69.5 & 16.0 & 18.4 & 32.4 & 26.0 & 33.5 \\
\midrule
\textbf{\name} & \textbf{74.4} & \textbf{91.0} & \textbf{64.0} & \textbf{80.3} & \textbf{82.3} & \textbf{60.5} & \textbf{75.4} \\ 
\bottomrule     
\end{tabular}
}
\caption{\textbf{Real-world action planning.} \name beats baselines on diverse zero-shot action planning. $\mathtt{CD, PF, SP, FS, RS}$ stands for \textit{Clear Desk}, \textit{Place Fruit}, \textit{Sort Parts}, \textit{Fold Shirts}, and \textit{Refill Shelf}, respectively. See \Cref{sec:planning_metric} for benchmark details.}
\label{tab:manipulation_results}
\end{table*}

\subsection{Action Planning}
\label{sec:planning_metric}

Real robot evaluation is time-consuming and entangles actor and planner failures. To cheaply and reliably evaluate planning performance we introduce an offline planning benchmark. The evaluation scores for this benchmark are given in \Cref{tab:manipulation_results}. \name significantly outperforms other models. Below  we provide details on how the planning benchmark is designed.

\textbf{Task Formulation.} We define our offline action planning as Multiple-Choice Questions (MCQ). At each decision point, the planner sees the current observation, semantic goal, and history, and selects the next subtask from dynamically generated candidates. Rather than scoring isolated frames, we simulate a continuous temporal rollout. Starting at $t=0$, we advance in fixed steps of size $\Delta t$. At each step, the current frame is extracted from the recorded video, the planner is queried, and the prediction is assigned to the interval $[t, t +\Delta t]$. This repeats until the episode ends. The memory harness injects sampled past images and prior predictions in a way that matches real-robot inference.                  
   
\textbf{Scoring Metrics.} We use temporal Intersection-over-Union rather than frame-wise accuracy. Each prediction is weighted by the duration of the ground-truth segment it overlaps. For each predicted segment we compute the overlap with ground-truth segments where the predicted action matches the ground truth label. Empirically, the planner ordering induced by this metric is consistent with the planner ordering on real-robot tasks.

\textbf{Benchmarks and Evaluation.} Our evaluation benchmark draws from two diverse and complementary data distributions to test both domain-specific robotic control and broad, out-of-distribution reasoning. The public \textbf{AgiBot dataset}~\cite{bu2025agibot} provides a rigorous testbed for standard robotic manipulation; it contributes five diverse task categories, each representing a distinct activity class: \textit{Clear Desk}, \textit{Place Fruit}, \textit{Sort Parts}, \textit{Fold Shirts}, and \textit{Refill Shelf}. Our internal \textbf{Egocentric Human-Hand dataset} pushes the boundaries of cross-embodiment generalization and open-world reasoning; it provides broad coverage of highly diverse, real-world human manipulation scenarios, encompassing 60 distinct tasks ranging from \textit{Organize Displays} and \textit{Make Spring Rolls} to \textit{Assemble Phones}, \textit{Carve Stones}, and \textit{Trim Rugs}. Each of these highly complex tasks is represented by a single, unique trajectory. 

Every episode within the benchmark features dense, step-level annotations. These include grounded, natural-language subtask instructions (\textit{e.g.}, "Pick up the green pear on the table with the right arm," "Place the green pear into the white floral-patterned plate") and precise frame-level temporal boundaries indicating the start and end of each subtask. Crucially, the subtask vocabulary is open-ended and dynamically defined per episode. The final evaluation suite is curated as a compact, highly challenging benchmark. It comprises 160 total episodes: 100 AgiBot trajectories (5 tasks $\times$ 20 episodes) and 60 Egocentric Human-Hand trajectories. All tasks included in this benchmark are strictly zero-shot; and are excluded from the training distribution.

\begin{figure}[!t]
\centering
\vspace{-3mm}
\begin{minipage}{0.58\linewidth}
\centering
\renewcommand{\arraystretch}{1.2}
\resizebox{\linewidth}{!}{
\begin{tabular}{l cccc}
\toprule
\textbf{Model} & $\mathtt{SR}\uparrow$ & $\mathtt{NE}\downarrow$ & $\mathtt{OS}\uparrow$ & $\mathtt{SPL}\uparrow$ \\
\midrule
RynnBrain-8B~\cite{dang2026rynnbrain}    & 0.0 & 8.86 & 0.0 & 0.0 \\
RoboBrain-2.5-8B~\cite{tan2026robobrain} & 0.0 & 9.03 & 0.0 & 0.0 \\
Qwen3-VL-8B~\cite{bai2025qwen3}        & 0.0 & 8.83 & 0.0 & 0.0 \\
UniNaVid~\cite{zhang2024uninavid}        & 47.0 & 5.58 & 53.3 & 42.7 \\
InternVLA-N1-8B~\cite{wang2025internvla} & \underline{55.4} & \textbf{4.89} & \underline{60.6} & \textbf{52.1} \\
\midrule
\textbf{\name}                           & \textbf{55.5} & \underline{5.16} & \textbf{61.4} & \underline{50.8} \\
\bottomrule
\end{tabular}
}
\captionof{table}{\textbf{Navigation in R2R-CE.} Navigation scores for \name and various baselines. Our model is on-par with the SOTA InternVLA-N1 navigation specialist model. It significantly beats all other generalist models.}
\label{tab:model_performance_reordered}
\end{minipage}
\hfill
\begin{minipage}{0.395\linewidth}
\centering
\includegraphics[width=\linewidth]{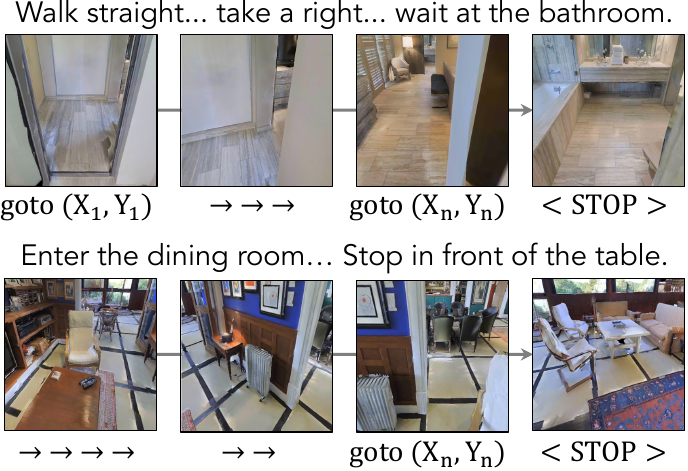}
\vspace{-4.8mm}
\captionof{figure}{\textbf{Demonstration of navigation evaluation}. Model outputs \textit{turns}, \textit{forward-to-locations}, and \textit{stop} actions. Intermediate steps are omitted.}
\label{fig:navigation_demonstration}
\end{minipage}
\end{figure}

\subsection{Navigation}

We evaluate navigation capabilities using the NavSuite benchmark. It runs the R2R \texttt{val\_unseen} split (1839 episodes in held-out Matterport3D scenes) inside the same Habitat simulator. Note that  all val-unseen scenes and episodes are excluded from our SFT data. At test time the agent's per-decision-point loop is fully closed: the planner is queried, its subgoal is dispatched to the low-level controller, and the agent's pose evolves until the planner emits \textsc{stop} or the 500-step budget is exhausted. We report the four standard R2R metrics: Success Rate (SR; fraction of episodes terminating within $d_\text{succ} = 3.0$~m of the goal), Success weighted by Path Length (SPL), Oracle Success (OS; success that would have been achieved had the agent stopped at the closest point along its trajectory), and Navigation Error (NE; geodesic distance between the final pose and the goal, in metres). All metrics are averaged over the 1839 val-unseen episodes. Scores are given in \Cref{tab:model_performance_reordered}. \name ties the SOTA navigation specialist InternVLA-N1 \cite{wang2025internvla}, leading on SR and OS while trailing on SPL and NE.

\begin{figure}[!t]
    \centering
    \vspace{-3mm}
    \includegraphics[width=\linewidth]{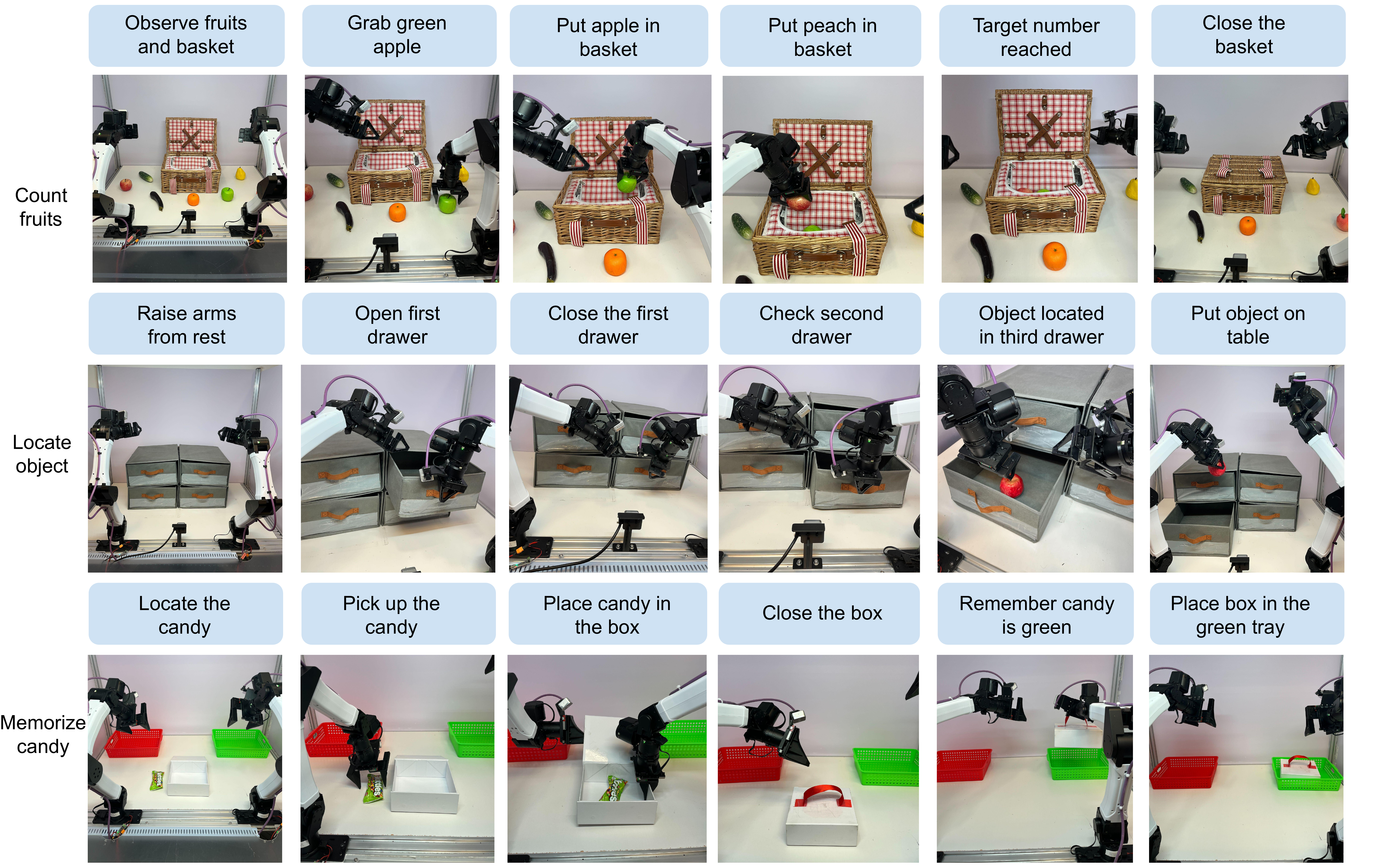}
    \caption{\textbf{Real robot tasks.} We evaluate \name as a planner model on real bimanual robots with three reasoning and memory-heavy tasks: find object, count fruits, and memorize candy.}
    \label{fig:real-evaluation-demo}
    \vspace{4mm}
\end{figure}

\begin{figure}[!ht]
    \centering
    \includegraphics[width=\linewidth]{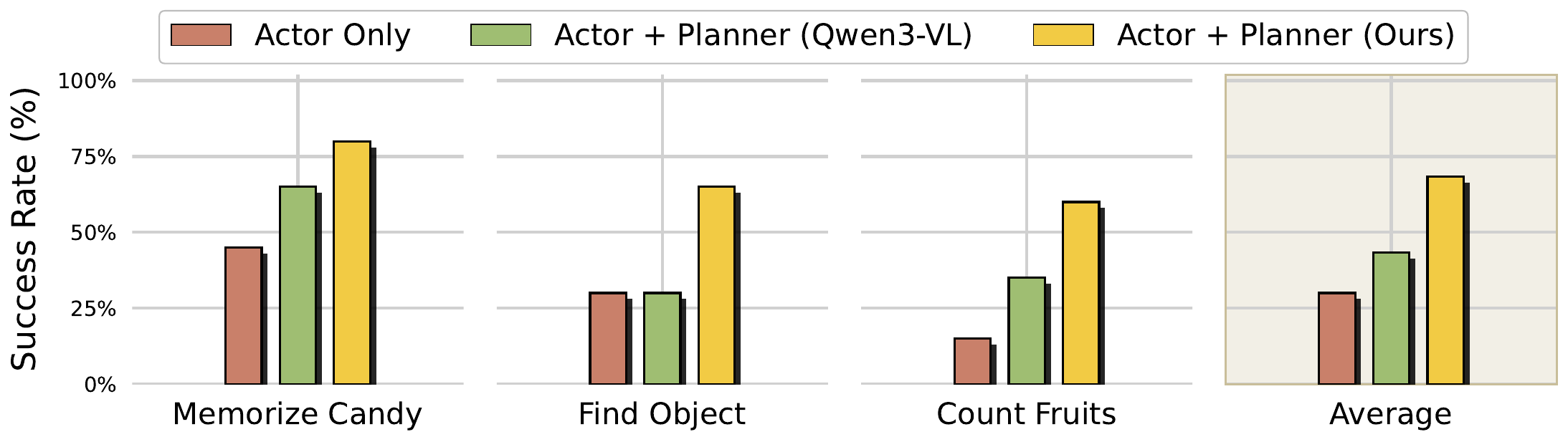}
    \caption{\textbf{Real robot evaluation}. Across tasks, \name significantly beats actor-only and Qwen3-VL baselines (w/ statistical significance over $4\sigma$). The average improvement over actor-only is 38.3\%.}
    \label{fig:real-evaluation}
\end{figure}

\subsection{Real Robot Evaluation} 

\label{sec:real_robot_eval}

We evaluate \name on real robotic manipulation tasks. We use the tabletop bimanual YAM grippers from I2RT robotics as the robotic platform. We consider the three following tasks:

\textbf{Find Object.} An object is placed in one of four compartments of a drawer. The task is to find the object by opening the drawers one-by-one, and then place it on the table. The task is terminated if the same drawer is opened twice. The planner thus needs to remember what drawers have been opened.

\textbf{Count Fruits.} A picnic basket and a number of fruits are placed on the table. The robot is instructed to place a specific number of fruits into the basket and then close it. The planner needs to instruct the actor to put the correct number of fruits into the basket one-by-one.

\textbf{Memorize Candy.} A box, a piece of candy and two colored trays are placed on the table. The candy should be placed in the box, the box closed, and then the box should be placed in the tray whose color matches the candy. Once the box is closed the planner needs to remember what it contains.

We use Gr00t-N1.6 \cite{bjorck2025gr00t} as the actor model. We evaluate three configurations: actor-only, the actor with a Qwen3-VL-8B~\cite{bai2025qwen3} planner, and the actor with a \name planner. Each task is evaluated with 20 samples; for hyperparameters and inference details see~\Cref{appendix:real_robot}. As shown in~\Cref{fig:real-evaluation}, using \name as the planner improves the average success rate by 38.3\% over the actor-only baseline, and by 25\% over the Qwen3-VL planner. With the given sample size, the $\sigma$ of the average is <9.2\%. So we beat the actor baseline with a statistical significance of $>4\sigma$. This demonstrates that \name can significantly improve real-robot execution and that our training significantly improves the base model. The actor itself will often make mistakes, either in motion quality or in language following. While the planner is not perfect, actor mistakes drive the majority of failed episodes. Due to constraints on our time with robots, don't evaluate specialist baselines optimized for academic benchmarks.

\subsection{Ablations}
\label{sec:ablations}

\begin{table}[!t]
\centering
\vspace{-3mm}
\resizebox{\linewidth}{!}{
\renewcommand{\arraystretch}{1.1}
\begin{tabular}{l|cc|cc|c}
\toprule
\multirow{2}{*}{\textbf{Training Mix}} & \multicolumn{2}{c|}{\textbf{Navigation (R2R-CE)}} & \multicolumn{2}{c|}{\textbf{Embodied}} & \multirow{2}{*}{\textbf{Avg.}} \\
 & SR $\uparrow$ & SPL $\uparrow$ & Cognition $\uparrow$ & Localization $\uparrow$ &  \\
\midrule
Nav-only specialist          & 54.1 & 49.8 & {0} & {0} & 26.0 \\
Embodied-only specialist     & 0 & 0 & {64.3} & {68.3} & {33.2} \\
\midrule
\rowcolor{gray!10}
\textbf{\name} (unified)     & {55.5} & {50.8} & \textbf{70.5} & \textbf{69.9} & \textbf{61.7} \\
\bottomrule
\end{tabular}
}
\caption{\textbf{Generalist vs.\ specialist training.} We fix other settings and only vary the data mixture. The unified model matches or beats each specialist in its domain, demonstrating positive transfer.}
\label{tab:general_training_ablation}
\vspace{8mm}
\end{table}

\textbf{Generalist vs. Specialist Training.} Unlike prior work that often trains separate specialists for navigation and embodied reasoning, we have shown that these capabilities can be combined into a unified model. Holding the architecture, base VLM, and total training budget fixed, we ablate only the data mixture: \textbf{Nav-only}, \textbf{Embodied-only}, and \textbf{\name} (unified mix). We evaluate all checkpoints on R2R val-unseen and the embodied benchmarks in \Cref{tab:embodied_benchmark}. As shown in \Cref{tab:general_training_ablation}, the unified model matches or outperforms each specialist on its own task (+1.4 SR on R2R; +3.9 avg. on embodied), indicating positive transfer and suggesting that a single generalist planner can not only match, but even outperform task-specific specialists.

\begin{wraptable}{rt}{0.49\textwidth}
  \centering
  \vspace{-4.0mm}
  \resizebox{0.49\textwidth}{!}{
    \renewcommand{\arraystretch}{1.1}
    \begin{tabular}{l ccc}
    \toprule
    \textbf{Setting} & \textbf{Trans.} & \textbf{Exec.} & \textbf{Overall} \\
    \midrule
    \multicolumn{4}{l}{\textit{Transition Sampling}} \\
    \midrule
    1$\times$ & 58.1 & \textbf{94.9} & 73.6 \\
    \rowcolor{gray!20} 2$\times$ & \underline{68.1} & \underline{90.0} & \textbf{75.9} \\
    3$\times$ & \textbf{69.1} & 88.7 & 75.3 \\
    \midrule
    \multicolumn{4}{l}{\textit{Memory Modality}} \\
    \midrule
    Image               & 61.0 & 70.4 & 63.1 \\
    Text                & 40.3 & 83.1 & 49.7 \\
    \rowcolor{gray!20} Image-Text Unif.    & \underline{68.1} & \textbf{90.0} & \textbf{75.9} \\
    Image-Text RecBias. & \textbf{68.9} & 87.5 & 75.3 \\
    \bottomrule
    \end{tabular}
  }
\vspace{-1mm}
\captionsetup{font=small}
\caption{\textbf{Ablation study.} We ablate oversampling ratio for transition-phase steps; and the modality of the history fed into the policy. Selected configurations are highlighted in gray.}
\label{tab:rl_ablation}
\end{wraptable}

\textbf{Transition sampling.} The transition phase, where the model must switch between actions, is far scarcer than the execution phase where the robot just continues the previous action (see \Cref{fig:nextaction_demonstration}). Under our default sampling frequency, transition-phase data accounts for only about $25\%$ of the total. As shown in~\Cref{tab:rl_ablation}, oversampling transition steps from $1\times$ to $2\times$ yields a large jump in transition accuracy and a clear gain in overall accuracy, while a further increase to $3\times$ brings only marginal improvement on transition and slightly hurts execution. We therefore adopt $2\times$ as our default.

\textbf{Memory design.} In \Cref{tab:rl_ablation} we ablate the modality (image-vs-text) of the memory design. \textit{Image-only memory} sacrifices accuracy: the model cannot easily understand its current progress from raw frames alone, and often decides to switch to a different action prematurely. \textit{Text-only memory} model instead learns to be overly reliant on the history text shortcuts, leading to excessive ``continue the current task'' predictions. Combining image and text memory strikes a balance and gives the best overall accuracy. We further compare \textit{uniform-} against \textit{recency-biased sampling} of frames: the two perform on par. This indicates that, conditioned on the right \emph{modality} mix (vision-text hybrid), the precise frame-selection policy is not the major bottleneck; Hence, a minimal design is sufficient in our setting.

\section{Related Work}

We provide a full literature review in \Cref{sec:appendix_related_work}. Here we provide an abridged version. 

\textbf{Embodied Vision-Language Models.} Using large pretrained vision language models as a "brain" for robots has been an active area of research since the inception of modern VLMs \cite{driess2023palm, huang2022inner}. A recent driving force has been the popularity of RL reasoning \cite{guo2025deepseek}, resulting in chain-of-thought becoming common methods for embodied control \cite{zawalski2024robotic}. The modern architecture that combines a planner VLM and an actor VLA has now become a standard method to introduce reasoning behavior into robotics \cite{shi2025hirobot,arxiv2025gpc}. For example, commercial VLAs now predict textual subtasks before executing actions \cite{intelligence2025pi05}. There is now a large group of VLMs specializing as the planners for embodied tasks \cite{tan2026robobrain,hao2026mimoembodiedxembodiedfoundationmodel,dang2026rynnbrain,ji2025robobrain, zhang2025pelican,team2025robobrain}, and there's growing evidence that generating points and traces can directly help low-level execution \cite{lee2025molmoact,zheng2025tracevla}. Many of these embodied VLMs are optimized for academic benchmarks rather than robust deployment. E.g. \citet{dang2026rynnbrain} provide multiple specialist finetuned models for different applications instead of a generalist checkpoint.

\textbf{Vision-Language Navigation.} Classical methods for robot navigation \cite{moravec1980obstacle} often use precomputed \cite{thrun1999minerva} or geometric maps that rely depth sensors \cite{newcombe2011kinectfusion} or monocular cameras that also localize the robot (SLAM) \cite{davison2007monoslam}. With the advent of foundation models, VLN models have dramatically improved by leveraging large pretrained models \cite{li2019robust}. Modern VLN specialists are thus typically obtained by finetuning VLMs \cite{cheng2024navila, wang2025internvla, zhang2025embodied} and are often evaluated in simulation environments \cite{vlnce,rxr}.

\textbf{Memory in Robotics.} Long-horizon embodied tasks require memory, yet many VLAs lack explicit history~\cite{brohan2023rt2, kim2024openvla, team2024octo, black2024pi0, bjorck2025gr00t, padalkar2023open, driess2023palme, wen2024tinyvla, wu2026pragmatic, ye2026world, doshi2024scaling}. Proposed compressions include spatial maps~\cite{henry2012rgbd, yu2024legs}, 2D visual traces~\cite{zheng2025tracevla, chen2025history, zhang2025tavla, chung2025rethinking, shi2025memoryvla, li2025cronusvla}, and keyframe retention~\cite{mark2026bpp, wei2025cycle, hu2025mllm, goletto2024amego, manigrasso2025online}; MemER uses experience retrieval to bound context~\cite{memer2026}, while MEM pairs short-horizon video with long-horizon language tracking~\cite{mem2026}, and others abstract events into language~\cite{lin2020image, sharma2023semantic, chiang2024mobility, chen2024commonsense, khazatsky2024droid, szot2024multimodal, zheng2025mllms} or split a high-level VLM from low-level control~\cite{shi2025hirobot, li2025hamster, shentu2024latent, intelligence2025pi05, zhao2023learning, wen2025dexvla, kamath2025gemma}.

\section{Discussion}

We present \name, a generalist embodied planner that matches or beats domain specialists on four capability axes simultaneously. Across the four capabilities, \name~on average scores >$20$ points above the strongest individual baseline and $>10$ points above an ensemble of baselines (i.e., taking the strongest model in each category). In real-world manipulation tasks, it increases success rate on memory-heavy tasks by 38.3\% over an actor-only baseline. Our work thus demonstrates that generalist planners are a feasible, scalable, and arguably preferable alternative to assembling specialists.

\bibliographystyle{plainnat}
\bibliography{main}

\newpage
\appendix
\section{Real Robot Evaluation}
\label{appendix:real_robot}

\begin{figure}[h]
    \centering
    \includegraphics[width=\linewidth]{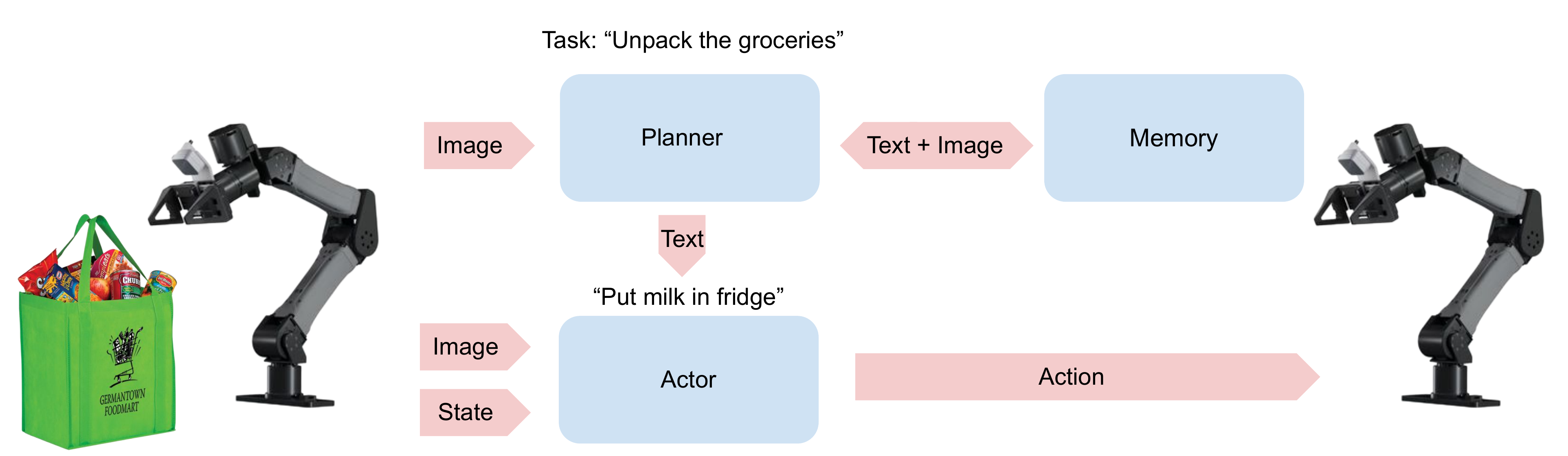}
    \caption{\textbf{Hierarchical execution.} The planner VLM takes as input a task specified in natural language, and continuously observes the images from the robot. It stores both text and image in a memory. At any given time, the planner sends the next subtask to the actor VLA. The actor consumes images, states and subtasks to produce robot actions. In this work, we focus on improving the planner.}
    \label{fig:s1s2}
\end{figure}

\begin{table}[h]
    \centering
    \renewcommand{\arraystretch}{1.35}
    \setlength{\tabcolsep}{6pt}
    {\renewcommand\tabularxcolumn[1]{m{#1}}%
    \begin{tabularx}{\linewidth}{@{}l c c >{\raggedright\arraybackslash}X@{}}
        \toprule
        \textbf{Task} & \textbf{\# Train} & \textbf{\# Eval} & \textbf{Success criterion} \\
        \midrule
        Count Fruits    & $401$ & $20$ & Basket closed, with the correct number of fruits inside, as specified by the task instruct. \\
        Find Object     & $774$ & $20$ & Object placed on the table with no drawer opened twice in the whole evaluation trajectory. \\
        Memorize Candy  & $500$ & $20$ & Candy placed in the box; box closed; and box placed on the correct tray with the same color as the candy. \\
        \bottomrule
    \end{tabularx}}
    \caption{\textbf{Robot task details.} Per-task sample counts and success criteria for the real-robot evaluations.}
    \vspace{4mm}
    \label{tab:robot_tasks}
\end{table}

Our robotic platform is the bimanual YAM grippers from I2RT robotics. For the real robot inference, we use the planner VLM to send text commands to the actor model. This follows the now standard setup \citep{shi2025hirobot}, and is illustrated in \Cref{fig:s1s2}. 

The detailed setup about the real robot tasks are given in \Cref{tab:robot_tasks}. We implement both synchronous and asynchronous coupling between the actor and the planner; both variants are described in \Cref{appendix:inference_loop}. For simplicity, all numbers reported in the main paper use the synchronous variant.

\section{Real-Robot Inference Loop}
\label{appendix:inference_loop}

\providecommand{\algorithmautorefname}{Algorithm}
\crefname{algorithm}{Algorithm}{Algorithms}
\Crefname{algorithm}{Algorithm}{Algorithms}

The real-robot deployment is a coupling between the high-level planner ($\pi_{\text{plan}}$, the \name~VLM) and a low-level actor ($\pi_{\text{act}}$, a Gr00t-N1.6 VLA~\cite{bjorck2025gr00t}). The two systems communicate through a single shared variable: the current natural-language \emph{subtask} $z$ (e.g.\ ``\texttt{pick up the green pear}''). The actor maps an environment observation $o_t$ together with $z$ to a low-level action $a_t$; it is not given the scene history or the planner's reasoning trace. The planner is a separate process that maps the high-level goal $g$, the latest observations $i_t$, and the memory $\mathcal{M}$; see \Cref{sec:methods}) to the next subtask $z'$. Memory is owned and managed by the planner server, and the actor never sees $\mathcal{M}$.

We expose two coupling modes between $\pi_{\text{plan}}$ and $\pi_{\text{act}}$, summarized side-by-side in \Cref{alg:sync_loop,alg:async_loop}. Both share the same control loop body and reset semantics; they differ only in \emph{when} a fresh planner inference is allowed to update $z$.

\begin{figure}[t]
    \centering
    \begin{minipage}[t]{0.48\textwidth}
        \captionof{algorithm}{Synchronous planner--actor loop.}
        \label{alg:sync_loop}
        \rule{\linewidth}{0.4pt}
        \begin{algorithmic}[1]
            \REQUIRE goal $g$, max staleness $\tau_{\max}$
            \STATE $z \gets \textsc{None}$,\ \ $t_{\text{last}} \gets -\infty$
            \STATE $\pi_{\text{plan}}.\textsc{ResetSession}()$
            \WHILE{episode not done}
                \STATE $o_t \gets \textsc{Env.Observe}()$
                \IF{$z = \textsc{None}$ \OR $g$ changed \OR $t - t_{\text{last}} \geq \tau_{\max}$}
                    \STATE $z \gets \pi_{\text{plan}}(i_t, g)$ \COMMENT{blocking}
                    \STATE $t_{\text{last}} \gets t$
                \ENDIF
                \STATE $a_t \gets \pi_{\text{act}}(o_t,\, z)$
                \STATE $\textsc{Env.Step}(a_t)$
            \ENDWHILE
        \end{algorithmic}
        \rule{\linewidth}{0.4pt}
    \end{minipage}\hfill
    \begin{minipage}[t]{0.48\textwidth}
        \captionof{algorithm}{Asynchronous planner--actor loop.}
        \label{alg:async_loop}
        \rule{\linewidth}{0.4pt}
        \begin{algorithmic}[1]
            \REQUIRE goal $g$, max staleness $\tau_{\max}$
            \STATE $z \gets \textsc{None}$,\ \ $t_{\text{last}} \gets -\infty$
            \STATE $\pi_{\text{plan}}.\textsc{ResetSession}()$
            \STATE \textbf{spawn} planner thread:
            \STATE \quad \textbf{loop:}
            \STATE \quad\quad $(i^\star,\, g) \gets \textsc{LatestObs}()$
            \STATE \quad\quad $z' \gets \pi_{\text{plan}}(i^\star,\, g)$
            \STATE \quad\quad atomically set $z \gets z'$, $t_{\text{last}} \gets t$
            \WHILE{episode not done}
                \STATE $o_t \gets \textsc{Env.Observe}()$;\ \ $\textsc{Publish}(i_t, g)$
                \IF{$z = \textsc{None}$ \OR $t - t_{\text{last}} \geq \tau_{\max}$}
                    \STATE \textbf{wait} for next planner publish
                \ENDIF
                \STATE $a_t \gets \pi_{\text{act}}(o_t,\, z)$
                \STATE $\textsc{Env.Step}(a_t)$
            \ENDWHILE
        \end{algorithmic}
        \rule{\linewidth}{0.4pt}
    \end{minipage}
\end{figure}

\textbf{How the two systems communicate.} In both modes the only inter-system traffic is (i) the actor publishing its latest observation to the planner, and (ii) the planner publishing its latest subtask to the actor; the planner's memory $\mathcal{M}$ is updated on the planner side at the end of each call and never crosses the boundary. The two modes implement the same freshness contract -- ``the actor never executes with a subtask older than $\tau_{\max}$'' -- but realize it differently. In the synchronous mode (\Cref{alg:sync_loop}), planner inference is a blocking call inside the control loop: the actor pauses for the duration of a planner step (typically $\sim$1\,s) whenever the goal changes or $\tau_{\max}$ has elapsed since the last subtask, and runs at full control rate in between. This is simple, deterministic, and gives the planner the freshest possible image at every replan event, at the cost of pauses in low-level motion. In the asynchronous mode (\Cref{alg:async_loop}), a dedicated planner thread runs continuously: as soon as one planner inference completes, the next one starts on the most recent published image. The control loop reads the cached subtask $z$ each tick without blocking and only stalls on the (rare) cold-start or cold-restart cases where no result has been produced within $\tau_{\max}$. The trade-off is that $z$ may correspond to an image that is up to one planner-inference old, but the actor runs uninterrupted at the full control rate. We use the synchronous variant for the numbers reported in the main paper because it is easier to log, reproduce, and ablate; the asynchronous variant is what we expect to use for production-style deployments where motion smoothness dominates.

\section{More on Training Details}
\label{appendix:training_details}

In this section we list the training details. The model SFT hyperparameters are given in \Cref{tab:sft_hps} and the separately fine-tuned Gr00t-N1.6~\cite{bjorck2025gr00t} actor used in the real-robot evaluation are given in \Cref{tab:vla_hps}. The actor and planner only interact at inference time through the natural-language subtask as described in \Cref{appendix:inference_loop}.

\textbf{Supervised Fine-Tuning.} We perform full-parameter SFT on Qwen3-VL-8B-Instruct~\cite{bai2025qwen3} using the data mixture described in \Cref{sec:training}. All three components of the model -- the vision tower, the multi-modal projector, and the language backbone -- are unfrozen and trained jointly in pure \texttt{bf16}. Optimization uses FSDP with full sharding across all data-parallel ranks (no parameter offload), FlashAttention-2 kernels, and sequence packing so that long-context multi-image samples can be batched without padding waste. The full set of SFT hyperparameters is summarized in \Cref{tab:sft_hps}.

\begin{table}[t]
    \centering
    \renewcommand{\arraystretch}{1.2}
    \setlength{\tabcolsep}{10pt}
    \resizebox{\linewidth}{!}{%
    \begin{tabular}{ll@{\hskip 24pt}ll}
        \toprule
        \textbf{Parameter} & \textbf{Value} & \textbf{Parameter} & \textbf{Value} \\
        \midrule
        Base model            & Qwen3-VL-8B-Instruct           & Per-device batch size  & $2$ \\
        Trainable parameters  & full (vision $+$ proj. $+$ LM) & Gradient accumulation  & $2$ \\
        Precision             & pure \texttt{bf16}             & Epochs                 & $1$ \\
        Learning rate         & $1\mathrm{e}{-5}$              & LR schedule            & cosine \\
        Weight decay          & $0$                            & Warmup ratio           & $0.1$ \\
        Max grad norm         & $1.0$                          & Sequence packing       & enabled \\
        Cutoff length         & $8{,}192$                      & Max video frames       & $32$ \\
        \midrule
        \multicolumn{4}{l}{\textbf{Infrastructure:}\quad H100 GPUs,\ \ FSDP full-shard (no offload),\ \ FlashAttention-2,\ \ gradient checkpointing} \\
        \bottomrule
    \end{tabular}}
    \caption{\textbf{SFT training details.} Hyperparameters used for the supervised fine-tuning stage of \name, starting from Qwen3-VL-8B-Instruct~\cite{bai2025qwen3}.}
    \vspace{5mm}
    \label{tab:sft_hps}
\end{table}

\textbf{Low-level actor VLA training.} For actor model, we adopt Gr00t-N1.6~\cite{bjorck2025gr00t}, which is a Vision-Language-Action (VLA) model. We train the VLA on collected demonstrations to ensure language following and motion smoothness. The VLA training hyperparameters are listed in \Cref{tab:vla_hps}.

\begin{table}[!ht]
    \centering
    \renewcommand{\arraystretch}{1.2}
    \setlength{\tabcolsep}{10pt}
    \begin{tabular}{ll@{\hskip 24pt}ll}
        \toprule
        \textbf{Parameter} & \textbf{Value} & \textbf{Parameter} & \textbf{Value} \\
        \midrule
        Base model     & Gr00t-N1.6           & Batch size      & $512$ \\
        Learning rate  & $1\mathrm{e}{-4}$    & Training steps  & $10{,}000$ \\
        Weight decay   & $1\mathrm{e}{-5}$    & Action horizon  & $50$ \\
        \midrule
        \multicolumn{4}{l}{\textbf{Color jitter:}\quad brightness $0.3$,\ \ contrast $0.4$,\ \ saturation $0.5$,\ \ hue $0.08$} \\
        \bottomrule
    \end{tabular}
    \caption{\textbf{Actor training details.} Hyperparameters used for fine-tuning the low-level actor policy.}
    \label{tab:vla_hps}
\end{table}

\section{Related Work}
\label{sec:appendix_related_work}

\paragraph{Embodied Foundation Models.} The pursuit of generalist embodied agents has driven rapid advancements in multimodal foundation models, successfully translating web-scale semantic knowledge into low-level robotic control and spatial reasoning~\cite{brohan2023rt2, google2025gemini, bai2025qwen3, team2024octo, kim2024openvla, black2024pi0, sapkota2025vla, ma2024vlasurvey, xu2025qwen25omni, sharshar2025omnibrain}. To handle the continuous dynamics of physical spaces, models have integrated specialized mechanisms such as continuous flow-matching~\cite{yuan2024generalflow, chen2025novaflow}, 3D representations~\cite{zhen20243dvla, lang2019pointpillars, liu2022bevfusion, chandorkar2025densepillarnet}, and state-space modeling~\cite{liu2024robomamba, abaza2025edgenavmamba, liu2024rdt1b, wang2024onedp, wu2025lightdp, arxiv2025flower, wang2025crisp}. 

Despite these structural leaps, models frequently struggle with multi-stage reasoning, cross-task adaptability, and long-horizon logic. Consequently, researchers increasingly explore hierarchical abstractions~\cite{chen2025fastinslow, park2025trace, li2025hamster, yue2024deervla}, explicit multimodal memory modules~\cite{gong2026acebrain, liu2025vlm2}, and post-training regimes (such as reinforcement learning and test-time composition)~\cite{dang2026rynnbrain, tan2026robobrain, ghasemipour2025selfimproving, ahn2023qtransformer, lin2025bfmzero, sun2025rlrc} to decouple fast, low-level execution from high-level cognitive planning~\cite{team2026gen1, xiang2025parallels, munje2026social, chen2025llarva, li2024robonurse, wang2024actra, ding2024quarvla, figureai2025helix, patel2025rigvid, mao2025physworld, zhang2024hover, zhao2025cotvla, duan2025fastecot, li2025cogvla, li2026qwen3vlembedding, cheng2024spatialrgpt, cai2025sensenovasi, yang2025visualspatialtuning, man2025argus, cvpr2025optimus1, cvpr20253dllmmem, cvpr2025sam2act, cvpr2025vagen, iclr2026vlm4vla, corl2025diwa, arxiv2025gpc, dong2025dreamvla}. However, prior works often treat this high-level policy as an isolated engineering component or rely on a patchwork of specialized networks. In contrast, our approach yields a unified, single generalist model capable of performing all reasoning, spatial, and cognitive QA tasks simultaneously. Rather than acting as a disconnected oracle, our model functions as a legitimate System-2 cognitive brain that explicitly manages dynamic context and memory, and is uniquely designed to be directly deployed to guide System-1 controllers via grounded language instructions in continuous real-robot execution.

\paragraph{Benchmarking and Evaluation of Robotics Foundation Models.} To systematically assess these cognitive capabilities, the community has witnessed a rapid increase of spatial intelligence and embodied question-answering benchmarks, rigorously testing visual perception, logical deduction, and spatial grounding across diverse egocentric and multi-view modalities~\cite{yang2025vsibench, yang2025mmsibench, gemini2025erqa, song2025robospatial, yin2025mindcube, wang2025metavqa, tong2024cvbench, fu2024blink, wei2025satbench, zhou2025roborefer, du2024embspatial, yuan2024robopoint, jia2025omnispatial, ji2025robobrain, yeh2025allangles, gholami2025ego3dbench, tang2025legopuzzle, oi2026hatch, han2020spare3d, liu2023vsr, kamath2023whatsup, zhang2025egothink, arxiv2025tracespatial}. Furthermore, diagnosing robotic failure modes, shortcut biases, and vulnerability to continuous environmental shifts has become essential for validating real-world robustness~\cite{liang2026robometer, liu2025robofac, wang2024vlbiasbench, narnaware2025sbbench, li2025aesbiasbench, shvetsova2025utdsplits, dunkel2025cnsbench}. While these datasets establish foundational metrics for static QA and spatial awareness, evaluating the complex, multi-stage decision-making of hierarchical policies traditionally requires prohibitively expensive online real-robot rollouts. To bridge this critical gap, we introduce a comprehensive, tri-fold evaluation suite: we establish a strong academic benchmarking baseline for SFT validation, propose a highly scalable offline evaluation framework sourced from diverse real and human hand datasets, and validate these proxy metrics against rigorous bimanual online real-robot evaluation, ensuring our System-2 model translates seamlessly from theoretical reasoning to physical action.

\paragraph{Memory in Robotics.} General-purpose embodied intelligence requires policies capable of acting in partially observable environments, making memory a critical component for resolving ambiguities and executing long-horizon tasks. While many recent VLA models operate purely reactively without explicitly managing historical context~\cite{brohan2023rt2, kim2024openvla, team2024octo, black2024pi0, bjorck2025gr00t, padalkar2023open, driess2023palme, wen2024tinyvla, wu2026pragmatic, ye2026world, doshi2024scaling}, scaling dense frame histories is computationally restrictive due to real-world latency constraints. To bypass these bottlenecks, various compression heuristics are proposed: spatial maps for navigation~\cite{henry2012rgbd, yu2024legs} relying purely on proprioceptive states, latent embeddings, and 2D visual traces for manipulation~\cite{zheng2025tracevla, chen2025history, zhang2025tavla, chung2025rethinking, shi2025memoryvla, li2025cronusvla}; or retaining only task-relevant keyframes ~\cite{mark2026bpp, wei2025cycle, hu2025mllm, goletto2024amego, manigrasso2025online}. Notably, the MemER framework utilizes experience retrieval by prompting a high-level policy to select informative keyframes to manage the context window without exceeding inference budgets~\cite{memer2026}.

Recognizing that no single modality perfectly captures both precise spatial requirements and high-level semantic progress, recent frameworks advocate for multimodal, multi-scale memory. For instance, MEM combines a short-horizon dense video encoder with a long-horizon language-based event tracker~\cite{mem2026}, while semantic memory approaches abstract past events into natural language to bypass visual processing costs~\cite{lin2020image, sharma2023semantic, chiang2024mobility, chen2024commonsense, khazatsky2024droid, szot2024multimodal, zheng2025mllms}. Hierarchical models that decompose reasoning into a high-level VLM and a low-level control policy have shown immense promise in managing these extended horizons~\cite{shi2025hirobot, li2025hamster, shentu2024latent, intelligence2025pi05, zhao2023learning, wen2025dexvla, kamath2025gemma}. Our approach is deliberately on the simple end of this spectrum: a fixed-budget sampler over recent image frames combined with a running text cache of prior subtasks, with no learned retriever or scene representation. We find this is sufficient to outperform single-modality variants of the same harness in our planner-actor setting (\Cref{sec:ablations}).

\section{Limitations}
\label{sec:limitations}

We highlight several limitations of \name~that we view as natural directions for future work.

\textbf{Real-robot evaluation scope.} Our online evaluation uses a single platform (the bimanual YAM grippers from I2RT robotics) with three memory- and reasoning-heavy tasks. The genuinely interesting next step is to move to embodiments where the planner-actor interface itself is stressed: humanoids that must couple whole-body locomotion with bimanual manipulation, mobile manipulators that switch between navigation and contact-rich phases mid-episode, and dexterous hands whose action vocabulary admits far finer subtasks than ``pick up the green pear''. Each of these breaks an implicit assumption of our current setup, namely that the actor can faithfully execute a small set of natural-language subtasks at a roughly fixed control rate. Studying how the planner's subtask abstraction should adapt to different actor controllability profiles is one of the central open questions in hierarchical embodied AI.

\textbf{Scaling, distillation, and edge deployment.} \name~is built on Qwen3-VL-8B~\cite{bai2025qwen3}, and all reported numbers are at the 8B scale. Most known scaling effects in VLMs have been characterized on web data, but embodied SFT mixtures are dominated by long-tail spatial and trajectory data whose scaling behavior is far less well understood.

\textbf{Learned and lifelong memory.} Our memory harness uses fixed retained frames with uniform or recency-biased sampling, plus a textual subtask log. One interesting next step is to make the memory itself an object of learning rather than a minimalist sampling policy. Lifelong settings beyond episodic memory (the same robot operating across many episodes in the same home or warehouse) introduce the additional question of when to consolidate, forget, or re-retrieve.

\section{Broader Impacts}
\label{sec:broader_impacts}

\name~targets generalist high-level planning for embodied agents and is, like much foundational ML research, several integration steps away from any specific deployment. We nonetheless surface impacts that we view as relevant.

\textbf{Potential positive impacts.} Generalist planners that subsume what is today a stack of brittle specialist models could meaningfully simplify deployment of assistive robots in domestic, healthcare, retail, and warehouse settings, and could lower the engineering barrier for academic and small-team robotics research. Reducing the number of separate models also reduces points of failure and makes overall system behavior easier to audit. The offline planning benchmark provides a low-cost ranking signal that can shrink the development cycle for groups that lack continuous access to physical platforms.

\textbf{Potential negative impacts.} Improved high-level planning, paired with capable low-level controllers, brings forward standard concerns about embodied AI: physical-safety risks from incorrect actions in shared spaces, possible labor displacement in routine physical tasks. Training and serving large VLM-based planners is also compute-intensive, which can concentrate capability in well-resourced organizations.

\textbf{Mitigation considerations.} The hierarchical planner/actor split that \name~uses is itself a useful safety surface: rule-based or learned safety filters can be applied at the planner output (textual subtasks) before any motor command is issued, and the actor can be sandboxed to a vetted action vocabulary. We restrict our own evaluation to a controlled lab environment with researcher supervision. When releasing assets in the future, we plan to accompany the release with documentation describing intended uses, evaluated capabilities and known failure modes, and recommended deployment guardrails.

\end{document}